\documentclass[letterpaper]{article} 
\usepackage{aaai2026}  
\usepackage{times}  
\usepackage{helvet}  
\usepackage{courier}  
\usepackage[hyphens]{url}  
\usepackage{graphicx} 
\usepackage{booktabs}
\usepackage{tabularx}
\usepackage{threeparttable}
\usepackage{amsmath}
\usepackage{multirow}
\usepackage{array}
\usepackage{natbib}  
\usepackage{caption} 
\usepackage{pgfplots}
\pgfplotsset{compat=1.18}
\usepackage{algorithm}
\usepackage{algorithmic}
\usepackage{newfloat}
\usepackage{listings}
\usepackage{float}
\usepackage{placeins}
\usepackage{caption}

\DeclareCaptionStyle{ruled}{labelfont=normalfont,labelsep=colon,strut=off}
\floatstyle{ruled}
\newfloat{listing}{tb}{lst}{}
\floatname{listing}{Listing}

\title{Are LLMs More Skeptical of Entertainment News?}
\author{Huiqian Lai}
\affiliations{
    School of Information Studies, Syracuse University\\
    hlai12@syr.edu
}

\begin{document}

\maketitle

\begin{abstract}
Large language models (LLMs) are increasingly used for automated news credibility assessment, yet it remains unclear whether they apply even-handed standards across journalistic genres. We examine whether zero-shot LLMs are more likely to misclassify legitimate entertainment news as fake than legitimate hard news, using a within-dataset design on GossipCop from FakeNewsNet. Across four frontier models, we find a clear but model-specific genre asymmetry: DeepSeek-V3.2 and GPT-5.2 show false-positive-rate gaps of 10.1 and 8.8 percentage points, respectively (both $p < .001$), whereas Claude Opus 4.6 and Gemini 3 Flash show no comparable difference. A style-swap experiment yields only limited and inconsistent changes, suggesting that the asymmetry is not reducible to stylistic register alone. Prompt-based mitigation is likewise possible but not generic: framing the model as an entertainment-news fact-checker reduces false positives for DeepSeek-V3.2 by about 50\% without detectable recall loss, but offers little improvement for GPT-5.2. Exploratory qualitative coding further suggests two recurring error patterns in sampled false positives: treating private-life claims as inherently unverifiable and discounting entertainment journalism as an epistemically weaker genre. Taken together, these findings show that aggregate performance metrics can obscure structured false positives within legitimate journalism. We argue that LLM-based credibility assessment may not only evaluate truth claims but also differentially recognize the legitimacy of journalistic genres, and that evaluation should therefore include genre-stratified false-positive analysis alongside overall accuracy.
\end{abstract}



\section{Introduction}

Entertainment journalism often reports on events that are later confirmed by mainstream outlets. For example, TMZ's 2009 report of Michael Jackson's death preceded network confirmation by nearly an hour~\citep{Guardian2009TMZJackson}, and the National Enquirer's coverage of John Edwards's extramarital affair broke a story that the mainstream press initially declined to pursue~\citep{CNN2008EdwardsCoverage}. These cases illustrate that entertainment reporting can produce factually accurate information, even when it operates outside traditional journalistic hierarchies.

At the same time, its stylistic conventions---such as a focus on private life and more narrative or emotionally vivid forms of presentation---overlap with surface-level features that stylometric and LLM-based detectors have been shown to rely on when assessing credibility~\citep{rashkin2017truth,potthast2018stylometric,horne2017thisjustin,wu2023sheepdog}. As large language models (LLMs) are increasingly used to evaluate news veracity at scale, this overlap raises a concrete concern: systems may misclassify legitimate entertainment journalism as unreliable. We use ``entertainment news'' and ``hard news'' in the sense established in journalism studies, and discuss this distinction in detail in \textit{Related Work}.


This concern points to a broader, structured risk in AI-mediated credibility assessment. Because automated systems often rely on stylistic and linguistic cues as proxies for truthfulness, they may disproportionately flag content that conforms to the conventions of particular journalistic genres~\citep{rashkin2017truth,horne2017thisjustin}. In the case of entertainment news, this creates the possibility of systematic false positives against legitimate reporting. Given the scale and visibility of entertainment news within contemporary media ecosystems, such asymmetries are not merely marginal but potentially consequential~\citep{reinemann2012hard, Turner}. They can bias platform moderation systems by disproportionately downranking certain types of content~\citep{gillespie2018custodians,roberts2019behind}, reinforce uneven recognition of journalistic legitimacy across genres~\citep{carlson2016metajournalistic,tong2018journalistic}, and ultimately erode user trust in AI-assisted information systems~\citep{sundar2020rise,flanagin2013trust}. Understanding whether and how such systematic misclassification arises is therefore critical for the responsible deployment of LLM-based credibility assessment~\citep{barocas2019fairness,mehrabi2021survey}.

Prior work on LLM-based misinformation detection has focused primarily on aggregate benchmark performance, prompting strategies, and domain generalization~\citep{chen2025explore,su2024adapting,gupta2025sok}. Related studies have also examined topic variation, domain-specific detection settings, and model vulnerability to stylistic perturbations~\citep{francis2024variation,cao2024scientific,wu2023sheepdog,tahmasebi2026adsent}. Yet this literature has not directly asked whether LLMs apply uneven credibility standards across journalistic genres, particularly by misclassifying legitimate entertainment news as fake. That omission matters because aggregate performance can mask systematic false positives against specific forms of legitimate journalism. We therefore still know little about whether entertainment news is disproportionately penalized, or whether any such pattern is driven by style, prompting, or broader assumptions about credibility.

We therefore ask the following research questions:

\noindent\textbf{RQ1:} Are zero-shot LLMs more likely to misclassify legitimate entertainment news as fake, compared to legitimate hard news?

\noindent\textbf{RQ2:} If such an asymmetry exists, is it driven primarily by stylistic cues, or by other aspects of how models evaluate credibility?

\noindent\textbf{RQ3:} Which prompt-based strategies can reduce these false positives, and what do recurring error patterns reveal about why they occur?

To answer these questions, we combine four steps: a within-dataset comparison, a style-swap test, a prompt-mitigation study, and qualitative error analysis, all based on GossipCop within FakeNewsNet~\citep{shu2020fakenewsnet}. This paper makes three contributions. \textit{First}, it identifies a model-specific genre asymmetry in zero-shot veracity classification: some frontier LLMs are substantially more likely to misclassify legitimate entertainment news as fake than legitimate hard news, even within the same dataset. \textit{Second}, it shows that this asymmetry is not reducible to stylistic register alone, as rewriting entertainment articles into a harder-news style produces only limited and inconsistent changes in model judgments. \textit{Third}, it demonstrates that mitigation is possible but not generic: an entertainment-domain expert prompt substantially reduces false positives for DeepSeek-V3.2 without detectable recall loss, but offers little improvement for GPT-5.2. More broadly, the study shows that aggregate performance metrics can obscure structured false positives within legitimate journalism, and that LLM-based credibility assessment may therefore not only evaluate truth claims but also differentially recognize which genres of journalism are presumptively trustworthy, effectively reproducing existing hierarchies of journalistic legitimacy.


\section{Related Work}
\subsection{Hard News, Soft News, and Entertainment News in Journalism Studies}
The distinction between hard news and soft news is one of the most enduring analytic categories in journalism studies~\citep{reinemann2012hard,LehmanWilzigSeletzky2010}. Hard news traditionally refers to reporting on public-affairs topics—politics, economics, and crime—characterized by high news value and temporal urgency~\citep{Tuchman,reinemann2012hard}. Soft news, by contrast, foregrounds lifestyle, personal experience, and emotional narrative, and includes entertainment and celebrity reporting as prominent examples~\citep{LehmanWilzigSeletzky2010}.

Scholarship has increasingly emphasized that this distinction is multidimensional rather than purely topical.~\citet{reinemann2012hard} synthesize prior work into three key dimensions—topic, focus, and style—arguing that news varies not only in subject matter but also in whether it emphasizes societal or individual relevance and whether it adopts impersonal or narrative-driven forms.~\citet[p.~51]{LehmanWilzigSeletzky2010} similarly critique the rigidity of the binary and propose an intermediate ``general news'' category for hybrid items that combine public-affairs content with softer presentation. Empirical work further shows that practitioners' judgments are shaped by presentational choices and role conceptions as much as by topic~\citep{Glogger10122019}. Beyond description, the distinction is embedded in a hierarchy of journalistic legitimacy, in which hard news has historically been treated as the prestige form of the profession, while soft and entertainment news are framed as lighter, more subjective, and less consequential~\citep{sjovaag2015hard}.

Despite this well-developed conceptual vocabulary, automated misinformation detection research has generally not engaged with it directly. Major surveys organize the field around content, style, propagation, and source credibility~\citep{Zhou2020}, and existing datasets typically treat entertainment as one topical domain among others rather than as a genre with distinctive evidentiary and stylistic conventions~\citep{perezrosas2018automatic,Silva_Luo_Karunasekera_Leckie_2021}. This leaves open a question that is central to the present study: whether the genre-level features used in journalism studies to characterize entertainment reporting—such as greater emphasis on individual relevance and more narrative or affective forms of presentation—are systematically interpreted by LLM-based credibility assessment systems as signals of unreliability.

\subsection{LLM-based Veracity Classification and Prompt Sensitivity}
Recent work on LLM-based misinformation detection suggests two related conclusions. First, advanced LLMs can serve as effective veracity classifiers and, in some settings, approach or exceed fine-tuned baselines~\citep{pelrine2023towards,vergho2024comparing}. Second, LLM judgments are not procedurally stable: semantically similar prompts, alternative evaluation formats, and other contextual framing cues can produce meaningful variation in outputs~\citep{SclarEtAl2024,zhuo2024prosa,germani2025source,errica2025sensitivity}. Taken together, this literature suggests that LLM-based credibility assessment is promising, but still highly sensitive to how the evaluative task is posed~\citep{SclarEtAl2024,zhuo2024prosa,germani2025source,vergho2024comparing}.

On the capability side,~\citet{pelrine2023towards} show that GPT-4 can outperform RoBERTa-large on LIAR and CT-FAN-22 in zero-shot settings, while exhibiting different failure modes from earlier supervised detectors.~\citet{vergho2024comparing} similarly find that performance varies substantially across models and prompt formulations, with notable instability across GPT-3.5 versions.~\citet{hu2024bad} add an important qualification: although GPT-3.5 underperforms fine-tuned BERT on direct veracity judgments, it can still provide informative multi-perspective rationales that improve downstream classification when incorporated into hybrid pipelines. At the same time, survey work positions LLMs as an increasingly important part of the fake-news detection landscape, especially for tasks such as classification, verification, and contextual analysis~\citep{chen2024combating,papageorgiou2024survey}. Relatedly,~\citet{leite2025pastel} show that LLM-extracted credibility signals can substantially improve article-level veracity classification, particularly in cross-domain settings, suggesting that LLM-based evaluation can be useful even when direct zero-shot judgments remain limited.

A growing body of research also shows that LLM evaluation is highly prompt-sensitive~\citep{SclarEtAl2024,zhuo2024prosa,germani2025source,errica2025sensitivity}.~\citet{SclarEtAl2024} demonstrate that meaning-preserving changes in prompt formatting can produce large swings in model performance.~\citet{zhuo2024prosa} similarly show that prompt sensitivity varies across tasks and becomes especially salient in subjective evaluation settings. Extending this concern beyond prompt wording alone,~\citet{germani2025source} find that LLM judgments shift systematically when source attribution is manipulated, indicating that evaluative outputs may depend not only on textual content but also on contextual framing cues surrounding that content.

What remains unclear, however, is whether these systems apply uneven credibility standards across different genres of legitimate journalism. Most existing studies evaluate overall benchmark performance, prompt robustness, or general detection capability~\citep{chen2024combating,papageorgiou2024survey,errica2025sensitivity,jin2025capefnd}. Even when they identify instability or bias, they do not directly test whether LLMs are more likely to misclassify legitimate articles from one journalistic genre as fake than equally legitimate articles from another. This omission matters because strong aggregate performance can obscure systematic false positives on particular forms of real journalism. We therefore still know little about whether zero-shot LLM credibility assessment is equally trustworthy across genres of legitimate news.

\subsection{Topic Domain Variation and the Neglect of Entertainment News}

Prior work suggests that entertainment news is not merely another topical category within fake-news detection, but a distinct detection domain with different linguistic cues, weaker cross-domain transfer, and different verification emphases~\citep{perezrosas2018automatic,silva2021embracing,mosallanezhad2022domain,shrestha2021textual}.~\citet{perezrosas2018automatic} introduced multi-domain fake news datasets spanning several topics and found clear domain variation in cross-domain evaluation. In their leave-one-domain-out experiments, politics was among the most robust domains, whereas entertainment was among the least generalizable, suggesting that signals learned in one domain do not transfer cleanly to another.~\citet{silva2021embracing} similarly compared PolitiFact and GossipCop and showed that differences in word usage and propagation patterns contribute to substantial performance degradation in cross-domain fake news detection.~\citet{mosallanezhad2022domain} further showed that domain-adaptive methods can improve target-domain performance even with limited target-domain supervision, confirming that the political-to-entertainment transfer gap is both real and learnable.

At the dataset level,~\citet{shu2020fakenewsnet} introduced FakeNewsNet, which includes PolitiFact and GossipCop as political and entertainment subsets, respectively, and explicitly highlighted the value of studying fake news across different news domains.~\citet{shrestha2021textual} performed systematic linguistic analysis across PolitiFact, BuzzFeedNews, and GossipCop and found that political and gossip news differ in their textual indicators of falsity. In particular, they show that fake political news contains more religion-related language, whereas gossip news relies more heavily on psychological and affective cues, and that feature usefulness varies across domains. Most directly,~\citet{li2024large} evaluated an agentic LLM detection system on both PolitiFact and GossipCop and designed its workflow so that a politics-oriented standing/bias tool is used only when the article is identified as political, indicating that verification priorities may need to vary by domain. Taken together, these studies show that entertainment news has long been present in benchmark design and evaluation, but mostly as a transfer domain or robustness test rather than as a primary object of analysis in its own right.

Together, these findings establish that entertainment news constitutes a genuinely distinct detection domain with its own linguistic norms, verification emphases, and failure modes. Yet, to our knowledge, the literature has rarely asked what this distinctiveness means for the treatment of legitimate entertainment journalism. Prior work uses GossipCop as a benchmark or transfer target, but it does not directly examine whether automated systems, and LLMs in particular, are more likely to distrust legitimate entertainment reporting than other forms of legitimate news. In other words, the missing question is not whether entertainment is a different domain, but whether its genre conventions are systematically misread as signs of falsity.

A plausible explanation, suggested by prior work on domain-specific linguistic cues and domain-sensitive verification workflows, is that automated detectors may over-rely on stylistic and tonal cues associated with entertainment news, treating them as signs of low credibility rather than as routine conventions of the genre.

\subsection{Stylistic Shortcuts and Their Genre-Conditioned Consequences}

A growing body of work suggests that misinformation detectors, whether fine-tuned neural models or prompted LLMs, often rely on linguistic form as a proxy for truthfulness rather than evaluating factual content directly~\citep{potthast2018stylometric,schuster2020limitations,wu2023sheepdog,wan2025truth,su2023biased}.~\citet{potthast2018stylometric} found that while stylometric features reliably distinguish hyperpartisan from mainstream content (F1\,=\,0.78), style-based fake news classification itself performed poorly (F1\,=\,0.46), suggesting that stylistic separability does not map cleanly onto veracity.~\citet{schuster2020limitations} extended this critique to machine-generated text, showing that stylometry can identify text provenance but cannot distinguish legitimate from misleading uses of language models.~\citet{wu2023sheepdog} further demonstrated that LLMs can camouflage fake news by rewriting it in the style of reputable outlets, causing up to a 38\% decline in F1 score in state-of-the-art detectors and confirming their vulnerability to stylistic surface features. More recently,~\citet{wan2025truth} introduced a systematic taxonomy of detector shortcuts spanning sentiment, style, topic, and perplexity, and showed that models degrade sharply under shortcut induction and injection, while also noting that real content may adopt subjective styles to increase engagement. A closely related finding comes from~\citet{su2023biased}, who show that fake-news detectors are biased against LLM-generated text: they flag LLM-generated fake news more readily than human-written fake news, and they also misclassify LLM-paraphrased real news as fake, which the authors interpret as evidence of detector reliance on shortcut-like linguistic cues. Related work on LLM-based credibility judgment further suggests that these evaluative systems may rely on lexical associations and statistical priors rather than contextual reasoning, and may confuse linguistic form with epistemic reliability~\citep{loru}.

These findings make style a plausible explanation for false positives on legitimate entertainment news, but not yet a sufficient one. Prior work has shown that detectors are vulnerable to stylistic manipulation, yet it has not directly tested whether rewriting legitimate entertainment news into a more conventional hard-news register reduces false positives, nor whether any remaining asymmetry reflects broader genre-level assumptions about credibility. The present study addresses this unresolved question by using a style-swap design to test whether style alone can account for LLM skepticism toward real entertainment news.

Taken together, these three strands of work motivate the present study. Prior research shows that LLM-based veracity assessment is increasingly capable but evaluatively unstable, that entertainment news constitutes a distinct detection domain, and that stylistic shortcuts may distort credibility judgments. What remains untested is whether these dynamics combine to produce systematic false positives on legitimate entertainment journalism, and whether any such asymmetry can be explained by style alone.

\section{Method}
\label{sec:method}
\subsection{Study Overview}
\label{subsec:overview}

We study whether zero-shot large language models (LLMs) treat legitimate entertainment news differently from legitimate hard-news reporting within the same dataset when asked to judge article veracity. Our design comprises three components. First, we compare false-positive rates on two groups of real articles within the same dataset to test for genre-conditioned bias while reducing dataset-level confounds. Second, we run a style-swap experiment in which real entertainment articles are rewritten into a hard-news style while keeping their core claims intact~\citep{ToshevskaGievska2021,ZhaoEtAl2024SC2}. Third, we test whether prompt-based mitigation can reduce false positives using an exploratory-to-confirmatory workflow~\citep{SclarEtAl2024}. We additionally conduct a qualitative error analysis to complement the quantitative findings. The unit of analysis in this study is the full news article, reflecting how credibility assessments---particularly in LLM-based settings---are typically conducted in practice. This approach also avoids the need to segment articles into discrete claims, a step that could introduce additional sources of variation~\citep{Lazer}.

This design is intentionally conservative. The evaluation is conducted on article text alone, without access to external verification signals such as sources, metadata, or retrieval. The primary comparison is conducted within GossipCop, a benchmark in the FakeNewsNet repository, so that the focal article groups are drawn from a shared collection environment rather than from different benchmark datasets~\citep{shu2020fakenewsnet}. As such, the design should be interpreted as a diagnostic test of model behavior rather than a fully identified causal estimate of the effect of stylistic features.

\subsection{Data}
\label{subsec:data}

We draw our data from FakeNewsNet, a benchmark collection of news articles labeled as real or fake~\citep{shu2020fakenewsnet}. Our primary analysis uses the GossipCop portion of FakeNewsNet as a controlled testbed in which articles of different genres coexist under a shared annotation and collection framework. Within GossipCop, we focus on real articles and compare two subsets: \textit{entertainment gossip} and \textit{hard news}.

\subsubsection{Operationalizing the two focal genres.}
Building on the hard/soft news distinction reviewed in Related Work, we operationalize the two focal categories along the topic, focus, and style dimensions identified by~\citet{reinemann2012hard}. An article is assigned to \textit{entertainment gossip} if it centers on celebrity, lifestyle, or personal-life matters and is presented in an emotionally vivid or narrative-driven register (e.g., affective language, dramatized framing, or anecdotal storytelling). An article is assigned to \textit{hard news} if it centers on public-affairs topics (e.g., politics, crime, major institutional events) and is presented in an event-centered, source-attributed register. Illustrative examples are provided in Appendix~A.5 (Table~\ref{tab:genre_examples}); detailed coding definitions are reported in Appendix~A.1--A.4.

\subsubsection{Labeling procedure and validation.}
Because GossipCop also contains articles that are neither pure entertainment gossip nor hard news---notably opinion commentary and promotional or PR-driven coverage---we use a four-way labeling scheme during the filtering stage so that these non-focal items can be separated out rather than misassigned to one of the two focal categories. Each article is classified into exactly one of four categories: \textit{entertainment gossip}, \textit{hard news}, \textit{opinion editorial}, or \textit{promotional}. We implement this step with DeepSeek-V3.2 (temperature = 0.0), using the article text as input and a single-label JSON output~\citep{deepseek_api_docs}. To validate the labels, we drew a stratified 10\% sample ($n = 400$) and had two human coders independently assign categories using the same coding instructions. Inter-coder agreement was high (Cohen's $\kappa = 0.86$), and agreement between the consensus human labels and the DeepSeek labels was 93\%. Three of the four evaluation models used in the main analysis (GPT-5.2, Claude Opus 4.6, Gemini 3 Flash) did not participate in this labeling step, which further reduces concerns about label circularity between genre assignment and veracity evaluation, as the labeling model is distinct from the evaluation models and the task of genre classification is orthogonal to veracity prediction.

\subsubsection{Evaluation subsets.}
After filtering, we retain only articles labeled as \textit{entertainment gossip} or \textit{hard news}; articles labeled as opinion or promotional are excluded from the main analysis. Focusing on real articles allows us to directly measure false-positive errors, which are central to our research question. Because all articles originate from GossipCop, the hard-news subset should be understood as a within-dataset reference group rather than a representative sample of institutional journalism. The main analysis uses 1,421 real entertainment-gossip articles and 379 real hard-news articles. The size imbalance between subsets reflects GossipCop's entertainment-focused construction. This limits the precision of the hard-news false-positive-rate estimate, but does not affect the direction of the comparison, which is the primary quantity of interest. The within-dataset design reduces the cross-dataset confounds that would arise in a direct comparison across different benchmarks. Additional evaluation-set details are reported in Appendix~A.6.

\subsection{Models and Inference Setup}
\label{subsec:models}

We evaluate four frontier models accessed through their official API identifiers: DeepSeek-V3.2~\citep{deepseek_api_docs}, GPT-5.2~\citep{openai_gpt52_docs}, Claude Opus 4.6~\citep{anthropic_models_overview}, and Gemini 3 Flash~\citep{google_gemini3flash_docs}. Temperature was fixed at 0.0 in all conditions, and maximum output length was capped at 512 tokens. For each article, the model was asked to return a JSON object containing a binary veracity judgment (\textit{real} or \textit{fake}), a credibility score between 0 and 1, and a one-sentence rationale. The baseline prompt was intentionally neutral and did not provide external evidence, retrieval support, or chain-of-thought instructions, so the setup approximates a zero-shot moderation or credibility-screening scenario; the full prompt wording is reported in Appendix~B.1.

\subsection{Experiment 1: Within-Dataset False-Positive Comparison}
\label{subsec:exp1}

The first experiment tests whether LLMs are more likely to produce false positives for legitimate entertainment news than for legitimate hard news. Using the same baseline prompt for all articles, we evaluate the two real-article subsets drawn from GossipCop under identical prompting conditions. For each model, we compute the false-positive rate (FPR) separately for real entertainment articles and real hard-news articles, where FPR is defined as the proportion of real articles predicted as \textit{fake}. We also record the model-generated credibility score for supplementary analysis.

Our main outcome is the difference in false-positive rates between the two groups:
\[
\Delta \mathrm{FPR}
=
\mathrm{FPR}_{\text{entertainment}}
-
\mathrm{FPR}_{\text{hard-news}}.
\]
A positive value indicates that the model is more likely to misclassify legitimate entertainment articles as fake than legitimate hard-news articles. We compare false-positive rates using two-proportion $z$-tests and report the difference in percentage points together with the associated $p$-value.

\subsection{Experiment 2: Style-Swap Test}
\label{subsec:exp2}

The second experiment examines whether the observed genre difference can be accounted for by stylistic form alone. We sample 50 real entertainment-gossip articles and use DeepSeek-V3.2 to rewrite each article into a more conventional hard-news style while keeping the core factual content as stable as possible. We then submit both the original article and its rewrite to all four evaluation models using the same baseline prompt.

Rewrite-fidelity checks indicated that the rewrites were adequate for diagnostic use; detailed automatic and manual fidelity results are reported in Appendix~A.7. For each model, we report aggregate fake rates before and after rewriting, together with paired correction and degradation rates. This experiment is not intended as a fully identified causal test of style, but as a diagnostic test of whether rewriting alone can systematically reduce false positives.

\subsection{Experiment 3: Prompt-Based Mitigation}
\label{subsec:exp3}

The third experiment tests whether prompting can reduce false positives on entertainment-news articles. We use an exploratory-to-confirmatory design with five prompt variants: a neutral baseline prompt (\textit{P0}), two lightweight heuristic prompts centered on verifiability and claim focus (\textit{P1}--\textit{P3}), and an expert-role prompt that frames the model as an entertainment-news fact-checker (\textit{P4}). Prompt definitions are summarized in Appendix~A.8 (Table~\ref{tab:app_prompt_strategy}); the full wording of the baseline and mitigation prompts is provided in Appendix~B.1--B.2.

The pilot phase used a pilot sample of 37 articles drawn from the false-positive cases identified for DeepSeek-V3.2 in Experiment~1 (approximately 15\% of the full DeepSeek-V3.2 false-positive pool of 249), together with a GPT-5.2 pilot set of 44 articles. These pilot analyses were used only to screen prompt variants and identify the most promising mitigation strategy; they are not treated as confirmatory evidence.

Based on the pilot results, we select P4 for confirmatory evaluation. In the confirmatory phase, we re-run all 249 DeepSeek false-positive cases identified in Experiment~1 using only P0 and P4. Both prompts are re-run within the same evaluation session to reduce the influence of API drift and session-level output instability. Because outputs may vary across sessions, the confirmatory correction rate is defined with respect to the subset of cases that remain false positives under the same-session P0 rerun:
\[
\mathrm{CorrectionRate}
=
\frac{N(\mathrm{P0}=\texttt{fake},\, \mathrm{P4}=\texttt{real})}
     {N(\mathrm{P0}=\texttt{fake})}.
\]
Under this definition, the denominator is 155 same-session baseline false positives rather than the original false-positive set from Experiment~1.

We also conduct a trade-off analysis on 347 fake entertainment articles to test whether P4 reduces false positives at the cost of lower recall on fake articles. We report correction rates with 95\% Wilson confidence intervals and use McNemar-style paired tests, with exact binomial variants where appropriate.

\subsection{Qualitative Error Analysis}
\label{subsec:qual}

To complement the quantitative results, we conduct a qualitative error analysis on 85 articles: 30 DeepSeek-V3.2 cases corrected by P4, 30 DeepSeek-V3.2 cases not corrected by P4, and 25 persistent GPT-5.2 false positives. For each case, we examine the article content, the model's rationale under the baseline prompt, and, where applicable, its rationale under mitigation prompting.

This analysis is not intended to estimate the prevalence of particular error types. Instead, it is used to interpret recurring forms of entertainment-news skepticism, including cases involving private-life unverifiability and cases in which entertainment journalism appears to be treated as an epistemically weaker genre.

\section{Results}

\subsection{Finding 1: Legitimate entertainment news is more likely to be misclassified as fake, but this pattern is model-specific.}
\label{subsec:find1}

\begin{figure}[t]
\centering
\begin{tikzpicture}
\begin{axis}[
    ybar,
    bar width=7pt,
    width=0.98\columnwidth,
    height=5.4cm,
    ymin=0,
    ymax=28,
    ylabel={False-positive rate (\%)},
    symbolic x coords={DeepSeek,GPT,Claude,Gemini},
    xtick=data,
    xticklabels={{DeepSeek\\V3.2},{GPT-5.2},{Claude\\Opus 4.6},{Gemini\\3 Flash}},
    xticklabel style={align=center, font=\scriptsize},
    ymajorgrids=true,
    grid style=dashed,
    enlarge x limits=0.18,
    legend style={
        font=\scriptsize,
        at={(0.5,1.02)},
        anchor=south,
        legend columns=1,
        draw=none
    }
]

\addplot+[
    error bars/.cd,
    y dir=both,
    y explicit,
] coordinates {
    (DeepSeek,17.5) +- (0,2.0)
    (GPT,20.9) +- (0,2.1)
    (Claude,1.5) +- (0,0.7)
    (Gemini,1.0) +- (0,0.6)
};

\addplot+[
    error bars/.cd,
    y dir=both,
    y explicit,
] coordinates {
    (DeepSeek,7.4) +- (0,2.6)
    (GPT,12.1) +- (0,3.3)
    (Claude,1.3) +- (0,1.5)
    (Gemini,0.8) +- (0,1.3)
};

\legend{Entertainment gossip ($n=1421$), Hard news ($n=379$)}
\end{axis}
\end{tikzpicture}
\caption{False-positive rates for real entertainment-gossip and real hard-news articles within GossipCop. Error bars indicate approximate 95\% confidence intervals. DeepSeek-V3.2 and GPT-5.2 show substantial genre-conditioned false-positive gaps, whereas Claude Opus 4.6 and Gemini 3 Flash do not.}
\label{fig:exp1_within_dataset_fpr}
\end{figure}
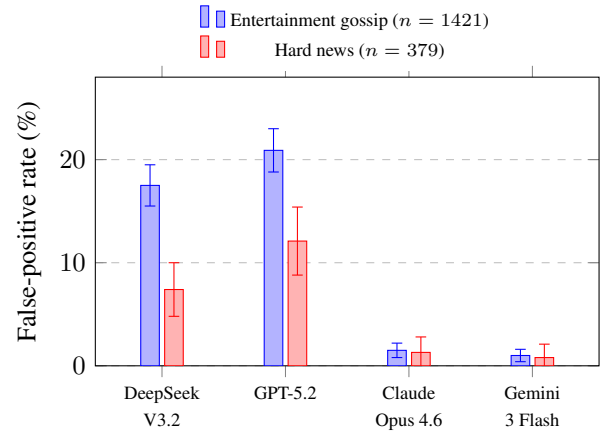

Our main within-dataset comparison shows that legitimate entertainment news is more likely to be misclassified as fake than legitimate hard news, but this pattern is not universal across models. As shown in Figure~\ref{fig:exp1_within_dataset_fpr}, the effect is concentrated in DeepSeek-V3.2 and GPT-5.2, whereas Claude Opus 4.6 and Gemini 3 Flash show little or no genre difference.

For DeepSeek-V3.2, the false-positive rate (FPR) on real entertainment-gossip articles is 17.5\% (249/1,421), compared with 7.4\% (28/379) on real hard-news articles, a difference of 10.1 percentage points (95\% CI [6.84, 13.43], $p<0.001$). GPT-5.2 shows a similarly large asymmetry, with an FPR of 20.9\% (297/1,421) on real entertainment articles and 12.1\% (46/379) on real hard-news articles, a difference of 8.8 percentage points (95\% CI [4.85, 12.67], $p<0.001$). By contrast, Claude Opus 4.6 and Gemini 3 Flash produce very low false-positive rates overall and show no statistically meaningful genre difference.

These null results matter substantively because they indicate that the entertainment-news penalty is not an inevitable property of zero-shot veracity classification, but a model-specific fairness concern in LLM-based credibility assessment. Exact test statistics are reported in Table~\ref{tab:app_find1_main}, and supplementary overlap and credibility-score analyses are reported in Appendix~C, including Table~\ref{tab:app_find1_credibility}.

\subsection{Finding 2: Style rewriting produces limited and inconsistent changes, which does not support a strong style-only explanation of the false-positive asymmetry.}
\label{subsec:find2}
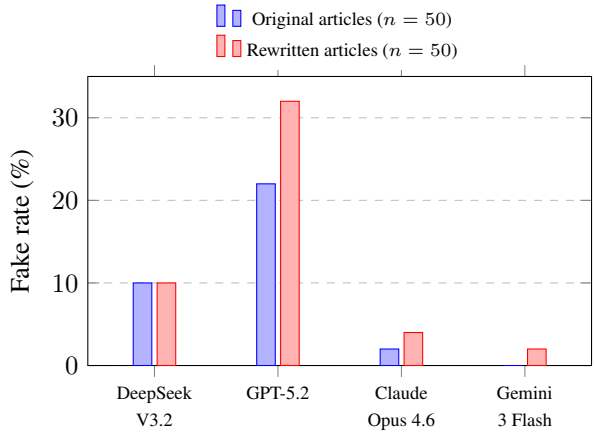
\begin{figure}[t]
\centering
\begin{tikzpicture}
\begin{axis}[
    ybar,
    bar width=7pt,
    width=0.98\columnwidth,
    height=5.4cm,
    ymin=0,
    ymax=35,
    ylabel={Fake rate (\%)},
    symbolic x coords={DeepSeek,GPT,Claude,Gemini},
    xtick=data,
    xticklabels={{DeepSeek\\V3.2},{GPT-5.2},{Claude\\Opus 4.6},{Gemini\\3 Flash}},
    xticklabel style={align=center, font=\scriptsize},
    ymajorgrids=true,
    grid style=dashed,
    enlarge x limits=0.18,
    legend style={
        font=\scriptsize,
        at={(0.5,1.02)},
        anchor=south,
        legend columns=1,
        draw=none
    }
]
\addplot coordinates {
    (DeepSeek,10.0)
    (GPT,22.0)
    (Claude,2.0)
    (Gemini,0.0)
};
\addplot coordinates {
    (DeepSeek,10.0)
    (GPT,32.0)
    (Claude,4.0)
    (Gemini,2.0)
};
\legend{Original articles ($n=50$), Rewritten articles ($n=50$)}
\end{axis}
\end{tikzpicture}
\caption{Fake rates before and after rewriting legitimate entertainment articles into a harder-news register. The dominant pattern is not systematic correction. DeepSeek-V3.2 shows no net change, GPT-5.2 becomes more skeptical after rewriting, and Claude Opus 4.6 and Gemini 3 Flash remain low-error overall. See Table~\ref{tab:app_exp2_style_swap_main} for raw counts and 95\% confidence intervals.}
\label{fig:exp2_style_swap_rates}
\end{figure}

The style-swap experiment provides little evidence that writing style alone explains the elevated false-positive rates observed for legitimate entertainment news. Across 50 paired cases, rewriting entertainment-gossip articles into a harder-news register produced only limited and inconsistent changes in model judgments. As shown in Figure~\ref{fig:exp2_style_swap_rates}, the dominant pattern is not systematic correction, but relative stability.

Rewrite-fidelity checks indicated that the rewritten articles were adequate for diagnostic use. The paired prediction results do not support a strong style-only explanation. For DeepSeek-V3.2, the aggregate fake rate remained unchanged at 10.0\% before and after rewriting, even though 3 of 5 original false positives were corrected and 3 originally correct judgments degraded. GPT-5.2 shows an even clearer mismatch with a style-only account: its fake rate increased from 22.0\% to 32.0\% after rewriting, with only 2 of 11 original false positives corrected and 7 of 39 originally correct judgments degrading into false positives.

The other two models again show that the pattern is not universal. Claude Opus 4.6 and Gemini 3 Flash remain low-error overall, and neither exhibits a systematic correction pattern after rewriting. Taken together, these results suggest that stylistic cues may matter at the margin, but they do not provide a sufficient explanation for the higher false-positive rates observed for legitimate entertainment news in Experiment~1. Full raw counts, confidence intervals, rewrite-fidelity checks, and illustrative paired examples are reported in Appendix D, including Table~\ref{tab:app_exp2_style_swap_main}.

\subsection{Finding 3: Prompt-based mitigation substantially reduces false positives for DeepSeek-V3.2, but remains weak for GPT-5.2.}
\label{subsec:find3}

The prompt-based mitigation results show that prompting can reduce false positives on legitimate entertainment news, but the effect is highly model-specific. Across the exploratory comparisons, simple verifiability reminders or claim-focus instructions do not reliably help. The only prompt variant that shows meaningful mitigation potential is P4, the expert-role prompt that frames the model as an entertainment-news fact-checker.

For GPT-5.2, the exploratory pilot provides little evidence that prompting offers a useful mitigation strategy. Across all four non-baseline prompts, fake-to-real flip rates remain low and the small pilot denominator yields wide confidence intervals. Even the best-performing variant, P4, corrects only 3 of 33 baseline false positives, while also introducing 7 real-to-fake reversals. Full GPT-5.2 pilot results are reported in Appendix~E, including Table~\ref{tab:app_exp3_gpt_pilot}.

DeepSeek-V3.2 behaves differently. In exploratory screening, P4 clearly outperforms the other prompt variants and is therefore selected for confirmatory evaluation. The main confirmatory test uses a same-session paired design on all 249 DeepSeek-V3.2 entertainment false positives identified in Experiment~1. This design is necessary because the baseline itself shows meaningful session-level instability even at temperature 0.0: 94 of the 249 articles (37.8\%) change their baseline label on rerun. We therefore treat same-session paired evaluation not only as a methodological control, but also as a more appropriate way to assess prompt effects under unstable API-based inference. For this reason, the confirmatory correction rate is defined over the 155 articles that remain false positives under the same-session P0 rerun.

A further trade-off analysis shows that this mitigation effect is not achieved simply by making the model more lenient overall. On a separate sample of 347 fake entertainment articles, DeepSeek-V3.2 attains a baseline fake-news recall of 51.3\% under P0 (95\% CI [46.0, 56.5]) and 53.9\% under P4 (95\% CI [48.6, 59.1]), a small positive difference of 2.6 percentage points. This change is not statistically significant ($p=0.289$), and the overlapping confidence intervals are consistent with the absence of a detectable recall trade-off. In other words, P4 reduces false positives on real entertainment articles without evidence of degraded fake-article detection.

Taken together, these results show that prompt-based mitigation is possible, but not generic. The intervention succeeds for DeepSeek-V3.2, but the same family of prompts yields only minimal improvement for GPT-5.2. Prompting is therefore best understood not as a universal fix, but as a model-specific interaction between prompt design and veracity reasoning.

\begin{table*}[t]
\centering
\begin{tabular}{@{}lcc@{}}
\toprule
Metric & Value & Interpretation \\
\midrule
Same-session baseline false positives & 155 & Confirmatory denominator \\
P4 corrections (F$\rightarrow$R) & 77 & Corrected by P4 \\
Confirmatory correction rate & 49.7\% [41.9, 57.5] & 77/155; 95\% Wilson CI \\
Paired exact test & $p<0.001$ & Significant improvement \\
\midrule
Fake-article sample size & 347 & Trade-off analysis \\
P0 fake recall & 51.3\% [46.0, 56.5] & Baseline recall on fake articles \\
P4 fake recall & 53.9\% [48.6, 59.1] & Recall under mitigation prompt \\
Recall difference & +2.6 pp & No evidence of degradation \\
Paired test on recall & $p=0.289$ & Not significant \\
\bottomrule
\end{tabular}
\caption{DeepSeek-V3.2 confirmatory mitigation and trade-off results for P4. The confirmatory correction rate is defined over the same-session P0 false positives only. Bracketed values indicate 95\% Wilson confidence intervals.}
\label{tab:exp3_deepseek_confirm}
\end{table*}

\subsection{Finding 4: Exploratory coding suggests that many sampled entertainment-news false positives reflect private-life unverifiability and genre-level distrust, but these patterns are not exhaustive.}
\label{subsec:find4}

The qualitative error analysis helps explain why the false-positive asymmetry persists after style rewriting and only partially responds to prompting. We conducted a lightweight exploratory coding of 85 entertainment-news false positives at the model--article level, including 60 cases from DeepSeek-V3.2 and 25 from GPT-5.2. Each case was assigned one primary error pattern after keyword-assisted pre-labeling and manual review. Rationales that did not clearly invoke either private-life unverifiability or genre-level distrust were coded as \textit{Other/unclear}.

Across the coded cases, 65.9\% exhibited one or both of the two hypothesized patterns. Private-life unverifiability was the most common single pattern, accounting for 29 of 85 cases (34.1\%), followed by genre-level distrust in 21 of 85 cases (24.7\%); 6 additional cases (7.1\%) showed both patterns. At the same time, 29 of 85 cases (34.1\%) remained in the \textit{Other/unclear} category, indicating that these two mechanisms explain a substantial share of false positives but do not exhaust the error space.

Substantively, these patterns help interpret the earlier experiments. Private-life unverifiability helps explain why style rewriting often yields limited gains, especially for stories whose evidentiary status is inherently indirect. Genre-level distrust helps explain why some models continue to reject entertainment reporting even when its presentation is rewritten into a harder-news register, and why prompting can reduce some errors without eliminating the asymmetry. Detailed pattern counts and illustrative cases are reported in Appendix E, including Table~\ref{tab:app_find4_error_patterns}.
\section{Discussion}
Across four frontier models and three complementary experiments, this study shows that zero-shot LLM-based veracity classification can apply uneven credibility standards within legitimate journalism.
More specifically, legitimate entertainment news is not judged neutrally by all models: DeepSeek-V3.2 and GPT-5.2 are significantly more likely to misclassify it as fake than legitimate hard news drawn from the same dataset, whereas Claude Opus 4.6 and Gemini 3 Flash show no comparable asymmetry.
The paper also clarifies the scope of that asymmetry.
It is not reducible to writing style alone, since rewriting entertainment articles into a more conventional hard-news register produces only limited and inconsistent changes, and in GPT-5.2 sometimes increases skepticism.
Nor is it uniformly correctable through prompting: an expert-role prompt substantially reduces DeepSeek-V3.2 false positives without degrading fake-article recall, but does not generalize to GPT-5.2.
Taken together, these results position entertainment-news skepticism as a model-specific failure mode in LLM credibility judgment, rather than an inherent property of zero-shot veracity classification itself.

\subsection{Empirical Implications}

Existing work on LLM-based veracity classification has largely emphasized benchmark performance, prompt robustness, and cross-domain generalization \citep{papageorgiou2024survey,errica2025sensitivity,leite2025pastel}. Our findings identify a different kind of failure: aggregate veracity performance does not guarantee even-handed credibility judgment across journalistic genres within legitimate news. In our within-corpus comparison of real articles, DeepSeek-V3.2 and GPT-5.2 were 10.1 and 8.8 percentage points more likely, respectively, to misclassify legitimate entertainment news as fake than legitimate hard news, whereas Claude Opus 4.6 and Gemini 3 Flash showed no comparable asymmetry. The empirical contribution, then, is to make visible a structured false-positive risk within legitimate journalism itself, one that conventional benchmark summaries are poorly equipped to capture.

Our experiments also narrow the plausible explanation for that failure mode. Prior work on prompt sensitivity and shortcut learning has shown that LLM judgments can vary under semantically equivalent prompts, contextual framing cues, and stylistic manipulation \citep{SclarEtAl2024,zhuo2024prosa,germani2025source,wan2025truth,wu2023sheepdog,su2023biased}. Our findings extend this line of work in two ways. First, they identify the complementary risk to style-based deception: not only can fake content be rewritten to look more credible, but legitimate journalism can also be misread as fake because its genre conventions resemble the surface features models associate with unreliability. Second, the experiments show that this asymmetry is neither reducible to stylistic register alone nor uniformly correctable through prompting. Rewriting entertainment articles into a more conventional hard-news register produces only limited and inconsistent changes, while expert-role prompting substantially reduces false positives for one model but fails to generalize to another. Taken together, these results suggest that the observed asymmetry reflects a deeper and model-specific form of skepticism, rather than prompt or register effects alone.

\subsection{Media-Theoretical Implications: Reproducing Journalistic Legitimacy Hierarchies}

Scholarship on journalistic legitimacy has generally treated legitimacy as the basis on which journalism secures cultural authority to produce valid knowledge of the world~\citep{carlson2016metajournalistic,tong2018journalistic,skovsgaard2011preference}. Carlson’s key intervention is to distinguish validity from truth: the issue is not only whether an account corresponds to reality, but whether particular forms and practices of knowing are recognized as legitimate in the first place~\citep{carlson2016metajournalistic}.~\citet{tong2018journalistic} likewise emphasizes that journalistic legitimacy is dynamic and must be continually maintained, while~\citet{skovsgaard2011preference} show why this problem is especially acute for journalism as a profession whose authority remains comparatively vulnerable and continuously in need of justification. 

Our findings extend this literature by showing that journalistic legitimacy may be unevenly allocated not only between journalism and its external challengers, but also across genres within journalism itself. In our results, some LLMs appear to grant legitimate entertainment reporting a weaker presumption of credibility than legitimate hard news, even when both are real articles. The implication is that these systems are not simply making isolated errors of factual classification; they are differentially recognizing what counts as a legitimate object of truth evaluation. Seen this way, the entertainment-news penalty is better understood as a model-specific downgrading of journalistic legitimacy across genres. More broadly, this suggests that LLM-based credibility assessment does not merely evaluate truth claims, but also implicitly ranks the legitimacy of different forms of journalism.

A more specific way to interpret this uneven allocation of credibility is through the hard/soft news hierarchy. The observed pattern---a weaker presumption of truth for legitimate entertainment news---is consistent with a long-standing hierarchy in journalism, in which hard news is treated as the most legitimate and authoritative form, while soft and entertainment news are framed as more emotional, subjective, and less consequential \citep{north2016gender,sjovaag2015hard,banjac2022struggle}. 

What is new in our findings is not the existence of this hierarchy, but its translation into AI-mediated credibility judgment. This suggests that LLMs are not neutral veracity evaluators, but systems that operationalize and reproduce existing genre hierarchies as epistemic criteria. In our data, this is most clearly reflected in the substantial false-positive gaps (approximately 10 percentage points) that DeepSeek-V3.2 and GPT-5.2 produce between legitimate entertainment and legitimate hard news---a gap that does not appear in Claude Opus 4.6 or Gemini 3 Flash.  This cross-model variation is itself diagnostic. Because the asymmetry is not universal across models, it is better understood not as an inherent property of entertainment news itself, but as a model-specific form of credibility sorting that reflects how different systems operationalize implicit assumptions about journalistic legitimacy. In this sense, variation across models is not simply noise, but evidence that these assumptions are contingent rather than fixed.

\subsection{Limitations}
Several limitations bound the scope of these conclusions. The primary analysis is confined to GossipCop within FakeNewsNet, so whether the observed entertainment-news penalty generalizes to other corpora, languages, platforms, or entertainment subgenres remains an open question. Within GossipCop, the hard-news subset ($n = 379$) is much smaller than the entertainment-gossip subset ($n = 1{,}421$), which limits the precision of hard-news false-positive estimates and means that this reference group should be understood as a within-corpus comparison category rather than a representative sample of institutional journalism more broadly. The evaluations are also text-only and were conducted under a fixed prompting protocol, so the results speak to how these models judge article text in isolation, not to how fuller verification systems might behave when given retrieval support, source metadata, publisher information, or social-context signals. The style-swap experiment is informative as a diagnostic, but it is not a fully identified causal test. Rewriting reduced article length by approximately 40\% and changed features beyond register alone, so the findings argue against a strong style-only explanation without isolating style as the sole causal factor. The qualitative error analysis was conducted by a single analyst on a convenience sample of 85 articles, and the coding categories were exploratory rather than theory-derived, so the reported proportions should be treated as suggestive patterns rather than estimates of the full false-positive population. Although all evaluations were conducted at temperature 0.0, API-based outputs were not perfectly stable across sessions: 94 of 249 DeepSeek-V3.2 baseline cases (37.8\%) changed labels on rerun. This degree of cross-call variation is itself important for interpreting the prompt-mitigation results, because it indicates that article-level judgments may shift even under nominally fixed inference settings. We therefore used a same-session paired design in Experiment 3 to reduce this source of variation. Even so, the remaining findings should be interpreted as snapshot measurements under a fixed evaluation protocol rather than as immutable properties of the underlying models.

Because the four models examined here represent a rapidly evolving technical landscape, both the magnitude of the genre asymmetry and the apparent effectiveness of prompt-based mitigation may shift as models, APIs, and deployment settings change over time.

 
\section{Conclusion}
 
This study shows that zero-shot LLMs can misrecognize legitimate journalism in genre-specific ways. In a within-dataset comparison on GossipCop, DeepSeek-V3.2 and GPT-5.2 were substantially more likely to label real entertainment news as fake than real hard news, whereas Claude Opus 4.6 and Gemini 3 Flash showed no comparable asymmetry. Across a style-swap test, prompt-based mitigation, and qualitative error analysis, we find that this pattern is not well explained by stylistic register alone. Instead, for some models, legitimate entertainment reporting appears to be granted a weaker presumption of credibility, especially when private-life claims are treated as inherently unverifiable or when entertainment journalism is implicitly discounted as a less trustworthy genre.

The broader implication is methodological as much as substantive. Aggregate benchmark performance can obscure structured false positives within real journalism, meaning that strong average results do not necessarily imply even-handed credibility judgment. For LLM-based misinformation detection, moderation, and news-assistance systems, evaluation should therefore include genre-stratified false-positive analysis rather than relying on overall accuracy alone. More broadly, the challenge is not only whether models can detect fake news, but whether they can recognize different genres of legitimate news as legitimate in the first place---and in doing so, avoid reproducing existing hierarchies of journalistic legitimacy.

\bibliography{aaai2026}

\clearpage
\FloatBarrier
\appendix
\onecolumn

\section*{Appendix A: Supplementary Materials for Method}

\subsection{A.1 LLM-Based Genre Classification}
\label{app:genre_llm}

To construct the within-GossipCop comparison subsets used in the main analysis, we classified articles into four genre categories using DeepSeek-V3.2 with temperature set to 0.0. This step was used only to assign genre labels for dataset construction. It was not used to generate article text.

For each article, we provided the article text to the model and asked it to assign exactly one of four labels: \textsc{HARD\_NEWS}, \textsc{ENTERTAINMENT\_GOSSIP}, \textsc{OPINION\_EDITORIAL}, or \textsc{PROMOTIONAL}. The model was instructed to return its output as a JSON object containing a genre label and a confidence score.

\subsection{A.2 Classification Prompt}
\label{app:genre_prompt}

We used the following prompt:

\begin{quote}
\small
\texttt{Classify this news article into ONE of four genres:}

\texttt{1. HARD\_NEWS - Factual, neutral tone; who/what/when/where/why structure; minimal evaluative language}

\texttt{2. ENTERTAINMENT\_GOSSIP - Exaggerated, emotionally vivid, hyperbolic; celebrity/personal focus; designed to entertain}

\texttt{3. OPINION\_EDITORIAL - Explicit subjective stance; passionate critique; persuasive intent}

\texttt{4. PROMOTIONAL - Reports on marketing/PR events; promotional language; brand partnerships}

\texttt{Article: \{text\}}

\texttt{Respond with ONLY a JSON object: \{"genre": "...", "confidence": 0.0-1.0\}}
\end{quote}

\subsection{A.3 Coding Instructions}
\label{app:genre_codebook}

The four genre categories were defined as follows, along with illustrative examples:

\begin{itemize}
    \item \textbf{ENTERTAINMENT\_GOSSIP}: Celebrity news, relationship updates, lifestyle coverage, and entertainment industry gossip. These articles typically foreground celebrity or personal matters and often use emotionally vivid, sensational, or hyperbolic language. \\
    \textit{Example}: ``Jennifer Aniston FINALLY breaks silence on Brad Pitt reunion rumors!''

    \item \textbf{HARD\_NEWS}: Articles oriented toward public-affairs topics such as politics, economics, science, crime, or major current events, and presented in a relatively more informational and event-centered register than celebrity gossip, opinion, or promotional content. \\
    \textit{Example}: ``Federal Reserve raises interest rates by 0.25 percentage points.''

    \item \textbf{OPINION\_EDITORIAL}: Opinion pieces, commentary, reviews, or analysis with an explicit subjective stance, evaluative framing, or persuasive intent. \\
    \textit{Example}: ``Why Hollywood's obsession with reboots is killing creativity.''

    \item \textbf{PROMOTIONAL}: Product announcements, sponsored content, press releases, or brand-partnership coverage characterized by promotional language or commercial intent. \\
    \textit{Example}: ``Kim Kardashian launches new SKIMS collection in collaboration with Fendi.''
\end{itemize}

\subsection{A.4 Validation Procedure}
\label{app:genre_validation}
To assess the quality of the LLM-generated genre labels, we manually validated a 10\% sample ($n=400$) of the classified articles. The validation sample was stratified by veracity label and included 200 real and 200 fake articles randomly sampled from the classified dataset.

Two human annotators independently labeled each article using the coding instructions above and without access to the model predictions. Inter-annotator reliability between the two human coders was high (Cohen's $\kappa = 0.86$). Disagreements were resolved through discussion to reach consensus. Agreement between the consensus human labels and the DeepSeek-generated labels was 93\% (372/400).

The main analysis retains only articles labeled as \textsc{ENTERTAINMENT\_GOSSIP} or \textsc{HARD\_NEWS}.

\subsection{A.5 Illustrative Examples of Focal Genre Categories}
\label{app:genre_examples}

\begin{center}
\begin{threeparttable}
\begin{tabularx}{\textwidth}{@{}p{4.0cm} X X@{}}
\toprule
\textbf{Genre} & \textbf{Example headline} & \textbf{Why it fits the label} \\
\midrule
\texttt{ENTERTAINMENT\_GOSSIP} & Katy Perry and Orlando Bloom Spotted Vacationing Together in the Maldives & Celebrity relationship coverage with soft, lifestyle-oriented gossip framing \\
\texttt{ENTERTAINMENT\_GOSSIP} & Brooke Burke Files for Divorce from David Charvet After Six Years & Celebrity private-life event framed in a tabloid-style register \\
\texttt{ENTERTAINMENT\_GOSSIP} & Meghan Markle and Prince Harry to Spend Night Apart Before Royal Wedding & Royal-wedding lifestyle detail framed as soft entertainment news \\
\texttt{HARD\_NEWS} & Oscars Debut New Rules To Avoid Another Envelope Mix-Up & Institutionally grounded coverage focused on industry rules and procedural reform \\
\texttt{HARD\_NEWS} & Toxicology Report Reveals High Fentanyl Concentration in Prince's Body & Investigative coverage centered on official forensic findings rather than gossip \\
\texttt{HARD\_NEWS} & Rose McGowan Faces Felony Arrest Warrant for Controlled Substance & Legal-process reporting focused on law-enforcement action and judicial procedure \\
\bottomrule
\end{tabularx}
\captionof{table}{Illustrative examples of the two focal genre categories used in the main analysis.}
\label{tab:genre_examples}
\end{threeparttable}
\end{center}

\subsection{A.6 Evaluation Sets}
\label{app:data_overview}

\begin{center}
\begin{threeparttable}
\begin{tabularx}{\textwidth}{@{}p{2.4cm} X p{1.3cm} p{3.0cm}@{}}
\toprule
\textbf{Component} & \textbf{Article set} & \textbf{$n$} & \textbf{Use} \\
\midrule
Exp.\ 1 (main) & GossipCop real entertainment gossip & 1,421 & Main false-positive comparison \\
Exp.\ 1 (main) & GossipCop real hard news & 379 & Main false-positive comparison \\
Exp.\ 2 & Real entertainment articles selected for style swap & 50 & Diagnostic style test \\
Exp.\ 3 pilot & DeepSeek pilot subset of baseline false positives & $\sim$15\% of pool & Prompt screening \\
Exp.\ 3 pilot & GPT pilot articles & 44 & Prompt screening \\
Exp.\ 3 conf. & DeepSeek false-positive pool from Exp.\ 1 & 249 & P0 vs.\ P4 evaluation \\
Exp.\ 3 recall & Entertainment fake articles for trade-off analysis & 347 & Recall trade-off \\
Qual.\ analysis & DeepSeek flipped, DeepSeek non-flipped, and GPT persistent false positives & 85 & Error/rationale analysis \\
\bottomrule
\end{tabularx}
\captionof{table}{Evaluation sets used in the study. The main comparison is within GossipCop.}
\label{tab:app_data_overview}
\begin{tablenotes}[flushleft]
\footnotesize
\item FP = false positive. In the confirmatory DeepSeek mitigation analysis, both P0 and P4 were re-run in the same session. The main correction-rate denominator is therefore the set of articles that remained false positives under the same-session P0 baseline, rather than the cached baseline from Experiment~1.
\end{tablenotes}
\end{threeparttable}
\end{center}

\subsection{A.7 Rewrite-Fidelity Checks for Experiment 2}
\label{app:exp2_fidelity}

Because Experiment~2 relies on rewritten versions of legitimate entertainment articles, we conducted both automatic and manual fidelity checks. The automatic checks yielded a mean TF--IDF cosine similarity of 0.5682 and a mean named-entity retention score of 0.5249 between each original article and its rewrite. The average article length decreased from 230.8 words in the original texts to 138.8 words after rewriting.

We also manually reviewed 10 rewritten articles. In 9 of the 10 cases, the rewrite preserved the key factual content sufficiently for diagnostic comparison. One case showed partial simplification of the original content, reinforcing the interpretation of Experiment~2 as a diagnostic test rather than a fully identified causal test of style alone.

\subsection{A.8 Prompt Strategy for Experiment 3}
\label{app:prompt_strategy}

\begin{center}
\begin{threeparttable}
\begin{tabularx}{\textwidth}{@{}p{1.2cm} p{5.4cm} p{2.4cm} X@{}}
\toprule
\textbf{Prompt} & \textbf{Idea} & \textbf{Stage} & \textbf{Purpose} \\
\midrule
P0 & Neutral zero-shot prompt & Pilot + conf. & Baseline reference \\
P1 & Verifiability reminder & Pilot only & Encourage separation of low visibility from falsity \\
P2 & Claim-focus prompt & Pilot only & Shift attention from genre tone to factual claims \\
P3 & Combined prompt & Pilot only & Test whether combining P1 and P2 improves performance \\
P4 & Expert entertainment fact-checker role & Pilot + conf. & Reframe the task with domain-specific expertise \\
\bottomrule
\end{tabularx}
\captionof{table}{Prompt strategy design for Experiment~3.}
\label{tab:app_prompt_strategy}
\begin{tablenotes}[flushleft]
\footnotesize
\item \textit{Note.} P1--P3 were tested in the pilot stage only. The full DeepSeek confirmatory run compares P0 and P4 only. GPT results are exploratory and remain pilot-only.
\end{tablenotes}
\end{threeparttable}
\end{center}

\section{B: Veracity Detection Prompts}
\label{app:veracity}

\subsection{B.1 Baseline Zero-Shot Veracity Prompt}
\label{app:veracity_prompt}

The baseline zero-shot prompt (P0) used for veracity detection across all models was:

\begin{verbatim}
Please read the following news article and determine
whether it is real or fake.

News text:
{text}

Respond with ONLY:
{"result": "real" or "fake",
 "credibility": 0.0-1.0,
 "rationale": "one sentence"}
\end{verbatim}

\subsection*{B.2 Mitigation Prompts}

Building on the baseline zero-shot veracity prompt reported in Appendix~B.1, we tested four prompt interventions designed to reduce genre-induced false positives on entertainment-news articles. Each variant preserves the same output format as P0 while modifying the framing of the evaluation task.

\subsubsection*{P1: Verifiability Reminder}

\begin{verbatim}
Please read the following news article and determine
whether it is real or fake.

Important: Entertainment and celebrity events are often
verifiable through public records, official statements,
social media, and credible news sources. Do not treat a
topic as unverifiable simply because it concerns personal
or celebrity matters.

News text:
{text}

Respond with ONLY:
{"result": "real" or "fake",
 "credibility": 0.0-1.0,
 "rationale": "one sentence"}
\end{verbatim}

\noindent\textit{Design rationale}: This prompt addresses the possibility that LLMs may implicitly treat entertainment topics as inherently less verifiable than other news topics.

\subsubsection*{P2: Explicit Claim Focus}

\begin{verbatim}
Please read the following news article and determine
whether it is real or fake.

First, identify the specific, verifiable factual claims
made in the article (e.g., names, dates, events, and
statements). Then assess only whether those claims are
accurate.

News text:
{text}

Respond with ONLY:
{"result": "real" or "fake",
 "credibility": 0.0-1.0,
 "rationale": "one sentence"}
\end{verbatim}

\noindent\textit{Design rationale}: This prompt encourages claim-level evaluation rather than a holistic judgment based on genre cues or writing style.

\subsubsection*{P3: Combined (P1 + P2)}

\begin{verbatim}
Please read the following news article and determine
whether it is real or fake.

Important: Entertainment and celebrity events are often
verifiable through public records, official statements,
social media, and credible news sources. Do not treat a
topic as unverifiable simply because it concerns personal
or celebrity matters.

First, identify the specific, verifiable factual claims
made in the article (e.g., names, dates, events, and
statements). Then assess only whether those claims are
accurate.

News text:
{text}

Respond with ONLY:
{"result": "real" or "fake",
 "credibility": 0.0-1.0,
 "rationale": "one sentence"}
\end{verbatim}

\noindent\textit{Design rationale}: This prompt tests whether combining a verifiability reminder with explicit claim extraction produces stronger mitigation than either intervention alone.

\subsubsection*{P4: Expert Role (Entertainment Fact-Checker)}

\begin{verbatim}
You are a fact-checker with expertise in entertainment
journalism. Assess the factual accuracy of the following
article based solely on its specific claims.

News text:
{text}

Respond with ONLY:
{"result": "real" or "fake",
 "credibility": 0.0-1.0,
 "rationale": "one sentence"}
\end{verbatim}

\noindent\textit{Design rationale}: This role-based prompt frames the model as a domain expert and tests whether an entertainment-specific fact-checking perspective reduces stylistic suspicion and improves veracity judgments.

\section{C: Supplementary Statistics for Finding 1}
\label{app:find1_stats}

Table~\ref{tab:app_find1_main} reports the full within-dataset statistics for Finding~1. The main result is that DeepSeek-V3.2 and GPT-5.2 show substantial false-positive asymmetries between real entertainment-gossip articles and real hard-news articles within GossipCop, whereas Claude Opus 4.6 and Gemini 3 Flash do not.

\begin{table}[H]
\centering
\begin{tabular}{lrrrrrl}
\toprule
Model & Ent. & Hard & $\Delta$ (pp) & 95\% CI & $z$ & $p$ \\
\midrule
DeepSeek-V3.2   & 17.5\% & 7.4\%  & +10.1 & [6.84, 13.43] & 4.86 & $<.001$ \\
GPT-5.2         & 20.9\% & 12.1\% & +8.8  & [4.85, 12.67] & 3.86 & $<.001$ \\
Claude Opus 4.6 & 1.5\%  & 1.3\%  & +0.2  & [-1.15, 1.47] & 0.23 & .82 \\
Gemini 3 Flash  & 1.0\%  & 0.8\%  & +0.2  & [-0.84, 1.22] & 0.35 & .73 \\
\bottomrule
\end{tabular}
\caption{False-positive rates on two real-article subsets from GossipCop: 1,421 entertainment-gossip articles and 379 hard-news articles. $\Delta$ indicates the entertainment-minus-hard-news difference in false-positive rate, reported in percentage points, with 95\% confidence intervals.}
\label{tab:app_find1_main}
\end{table}

We also examined whether the two affected models failed on the same entertainment articles. DeepSeek-V3.2 produced 249 entertainment false positives and GPT-5.2 produced 297, with an overlap of 135 articles, corresponding to a Jaccard similarity of 0.33. This partial overlap suggests that the entertainment-news asymmetry is shared across the two affected models, while the exact boundary of suspicious entertainment content remains model-dependent.

A supplementary analysis of continuous credibility scores showed the same directional pattern. On real articles, DeepSeek-V3.2 assigned slightly lower average credibility to entertainment than to hard news (0.940 vs.\ 0.951), and GPT-5.2 showed a larger gap (0.785 vs.\ 0.848), as shown in Table~\ref{tab:app_find1_credibility}. These supplementary results provide convergent support for the main finding that some models are systematically more skeptical of legitimate entertainment news than of legitimate hard news within the same dataset context.

\begin{table}[H]
\centering
\begin{tabular}{lcc}
\toprule
Model & Entertainment & Hard news \\
\midrule
DeepSeek-V3.2 & 0.940 & 0.951 \\
GPT-5.2 & 0.785 & 0.848 \\
\bottomrule
\end{tabular}
\caption{Average credibility scores on real entertainment-gossip and real hard-news articles for the two models that show a substantial false-positive asymmetry in the main analysis.}
\label{tab:app_find1_credibility}
\end{table}
\section{D: Supplementary Materials for Finding 2}
\label{app:find2_supp}

Table~\ref{tab:app_exp2_style_swap_main} reports the full style-swap results for 50 paired real entertainment articles.

\begin{table}[H]
\centering
\begin{tabular}{lcccc}
\toprule
\textbf{Model} & \textbf{Original fake rate} & \textbf{Rewrite fake rate} & \textbf{Correction rate} & \textbf{Degradation rate} \\
\midrule
DeepSeek-V3.2 
& 10.0\% (5/50) [4.3, 21.4] 
& 10.0\% (5/50) [4.3, 21.4] 
& 60.0\% (3/5) [23.1, 88.2] 
& 6.7\% (3/45) [2.3, 17.9] \\

GPT-5.2 
& 22.0\% (11/50) [12.7, 35.3] 
& 32.0\% (16/50) [20.8, 45.8] 
& 18.2\% (2/11) [5.2, 47.7] 
& 17.9\% (7/39) [8.9, 32.6] \\

Claude Opus 4.6 
& 2.0\% (1/50) [0.4, 10.5] 
& 4.0\% (2/50) [1.1, 13.5] 
& 100.0\% (1/1) [20.7, 100.0] 
& 4.1\% (2/49) [1.1, 13.7] \\

Gemini 3 Flash 
& 0.0\% (0/50) [0.0, 7.1] 
& 2.0\% (1/50) [0.4, 10.5] 
& N/A (0/0) 
& 2.0\% (1/50) [0.4, 10.5] \\
\bottomrule
\end{tabular}
\caption{Style-swap results for 50 paired real entertainment articles. Correction rate is the proportion of original false positives that flipped from \textit{fake} to \textit{real} after rewriting. Degradation rate is the proportion of original correct predictions that flipped from \textit{real} to \textit{fake}. Values are reported as percentages with raw counts in parentheses; bracketed values indicate 95\% confidence intervals. Because some correction-rate denominators are very small, several interval estimates are necessarily wide.}
\label{tab:app_exp2_style_swap_main}
\end{table}

We also evaluated whether the rewrites were sufficiently faithful for diagnostic comparison. On average, article length decreased from 230.8 words to 138.8 words, corresponding to a 39.9\% reduction. The mean TF--IDF cosine similarity between each original article and its rewrite was 0.568, and mean named-entity retention was 0.525. In a manual review of 10 sampled pairs, 9 rewrites were judged to preserve the key factual content sufficiently for analysis. One partial case involved a fashion commentary article with minimal verifiable factual content, reinforcing the interpretation of Experiment~2 as a diagnostic test rather than a fully identified causal test.

Two illustrative paired cases help clarify the aggregate pattern. In \textit{gossipcop-907697}, a report about legal filings involving the Kardashian family, GPT-5.2 labeled both the original entertainment-style version and the harder-news rewrite as \textit{fake}. By contrast, in \textit{gossipcop-855820}, DeepSeek-V3.2 changed from \textit{fake} on the original article to \textit{real} on the rewritten version. These examples are consistent with the broader quantitative result: rewriting can occasionally shift predictions, but it does not reliably eliminate the underlying asymmetry.

\section{E: Supplementary Materials for Finding 3}
\label{app:find3_supp}

Table~\ref{tab:app_exp3_gpt_pilot} reports the exploratory GPT-5.2 pilot results across four mitigation prompts. Across all four non-baseline prompts, fake-to-real flip rates remain low and the net error profile does not improve.

\begin{table}[H]
\centering
\begin{tabular}{@{}lcccc@{}}
\toprule
Prompt & $n$ & F$\rightarrow$R flips & Flip rate (95\% CI) & Notes \\
\midrule
P1 & 33 & 2 & 6.1\% [1.7, 19.6] & 8 R$\rightarrow$F; net worse \\
P2 & 33 & 0 & 0.0\% [0.0, 10.4] & 3 R$\rightarrow$F; net worse \\
P3 & 33 & 1 & 3.0\% [0.5, 15.3] & 7 R$\rightarrow$F; net worse \\
P4 & 33 & 3 & 9.1\% [3.2, 23.6] & 7 R$\rightarrow$F; net worse \\
\bottomrule
\end{tabular}
\caption{Exploratory GPT-5.2 pilot results across four mitigation prompts. Flip rates are calculated over baseline false positives in the pilot set. Bracketed values indicate 95\% Wilson confidence intervals.}
\label{tab:app_exp3_gpt_pilot}
\end{table}

\section{E: Supplementary Materials for Finding 4}
\label{app:find4_supp}

Table~\ref{tab:app_find4_error_patterns} reports the distribution of exploratory error-pattern labels across 85 entertainment-news false positives. The coding was conducted at the model--article level on a convenience sample, including 60 cases from DeepSeek-V3.2 and 25 from GPT-5.2. Each case was assigned one primary label after keyword-assisted pre-labeling and manual case-by-case review: \textit{private-life unverifiability}, \textit{genre-level distrust}, \textit{both}, or \textit{other/unclear}. Rationales that did not clearly invoke either of the two focal mechanisms were coded as \textit{Other/unclear}.

\begin{table}[h]
\centering
\small
\setlength{\tabcolsep}{2mm}
\renewcommand{\arraystretch}{1.08}
\begin{tabular}{@{}lcccc@{}}
\toprule
\multirow{2}{*}{Error Pattern} & \multicolumn{2}{c}{DeepSeek-V3.2} & \multicolumn{2}{c}{GPT-5.2} \\
\cmidrule(lr){2-3} \cmidrule(lr){4-5}
 & $n$ & \% & $n$ & \% \\
\midrule
Private-life unverifiability & 20 & 33.3\% & 9 & 36.0\% \\
Genre-level distrust & 16 & 26.7\% & 5 & 20.0\% \\
Both patterns & 5 & 8.3\% & 1 & 4.0\% \\
Other/unclear & 19 & 31.7\% & 10 & 40.0\% \\
\midrule
Total & 60 & 100\% & 25 & 100\% \\
\bottomrule
\end{tabular}
\caption{Distribution of coded error patterns in entertainment-news false positives ($N=85$). Labels were assigned via keyword-assisted pre-labeling followed by manual case-by-case review.}
\label{tab:app_find4_error_patterns}
\end{table}

The first recurring pattern, private-life unverifiability, is most visible in stories involving romantic relationships, family tensions, injuries, health, or other off-stage personal events. In these cases, the model often treats claims as suspicious because they are difficult to verify through public institutional evidence, even when such claims are commonly reported through interviews, representatives, court filings, televised appearances, or entertainment media outlets.

The second recurring pattern, genre-level distrust, is more structural. In these cases, the model appears to treat entertainment journalism itself as an epistemically weak domain. This pattern is especially visible when a change in stylistic presentation fails to alter the judgment.

Two illustrative paired cases help clarify the aggregate pattern. In \textit{gossipcop-907697}, a report about legal filings involving the Kardashian family, GPT-5.2 labeled both the original entertainment-style version and the harder-news rewrite as \textit{fake}. The rewrite removed much of the entertainment framing and restated the story in a more neutral reportorial style, yet the model's judgment did not change. By contrast, in \textit{gossipcop-855820}, DeepSeek-V3.2 changed from \textit{fake} on the original entertainment-style article to \textit{real} on the rewritten version. Together, these cases suggest that stylistic cues can occasionally shift decisions, but that some errors are tied to deeper assumptions about the verifiability of private-life claims or the epistemic status of entertainment reporting.

These patterns also help interpret the asymmetric prompt results in Experiment~3. The P4 mitigation prompt appears most effective when the error is driven by a relatively shallow heuristic and the model can be redirected toward evaluating a concrete claim. Prompting is less effective when the story concerns intimate personal life, ambiguous relationship dynamics, or events whose evidentiary status is intrinsically indirect. In such cases, the model's skepticism appears to stem not only from genre cues, but from a deeper mismatch between entertainment reporting and the model's implicit assumptions about reliable evidence.

\end{document}